\pdfoutput=1

\documentclass[11pt]{article}

\usepackage{coling}
\usepackage{todonotes}

\usepackage{times}
\usepackage{latexsym}
\usepackage{color}
\usepackage{tikz}
\usepackage{CJKutf8}
\usepackage{xcolor}
\usepackage{microtype}
\usepackage{mathtools}
\usepackage{adjustbox}
\usepackage{enumitem}
\usepackage{tablefootnote}
\usepackage{multirow}
\usepackage{bm}
\usepackage{amsmath}
\usepackage{amssymb}
\usepackage{booktabs}
\usepackage{ulem}

\usepackage{graphicx}

\usepackage[T1]{fontenc}

\usepackage[utf8]{inputenc}

\usepackage{microtype}

\usepackage{inconsolata}

\title{Mining Word Boundaries from Speech-Text Parallel Data for \\ Cross-domain Chinese Word Segmentation} 

\author{
 \textbf{Xuebin Wang},
 \textbf{Lei Zhang},
 \textbf{Zhenghua Li\thanks{Corresponding author}},
 \textbf{Shilin Zhou},
 \textbf{Chen Gong},
 \textbf{Yang Hou}
\\
 School of Computer Science and Technology, Soochow University, China \\
\texttt{\{xbwang15,yhou1\}@stu.suda.edu.cn}, 
\texttt{\{zhli13,gongchen18\}@suda.edu.cn}, \\
\texttt{leizhang.nlp@gmail.com}, 
\texttt{slzhou.cs@outlook.com},
}

\begin{document}
\maketitle
\begin{CJK}{UTF8}{gkai}

\begin{abstract}

Inspired by early research on exploring naturally annotated data for Chinese Word Segmentation (CWS), and also by recent research on integration of speech and text processing, this work for the first time proposes to explicitly mine word boundaries from speech-text parallel data. 
We employ the Montreal Forced Aligner (MFA) toolkit to perform character-level alignment on speech-text data, giving pauses as candidate word boundaries. 
Based on detailed analysis of collected pauses, we propose an effective probability-based strategy for filtering unreliable word boundaries. 
To more effectively utilize word boundaries as extra training data, we also propose a robust complete-then-train (CTT) strategy. 
We conduct cross-domain CWS experiments on two target domains, i.e., ZX and AISHELL2. 
We have annotated about 1,000 sentences as the evaluation data of AISHELL2. 
Experiments demonstrate the effectiveness of our proposed approach.





\end{abstract}

\section{Introduction}

\begin{figure}[tb]
\centering
\includegraphics[width=7.5cm]{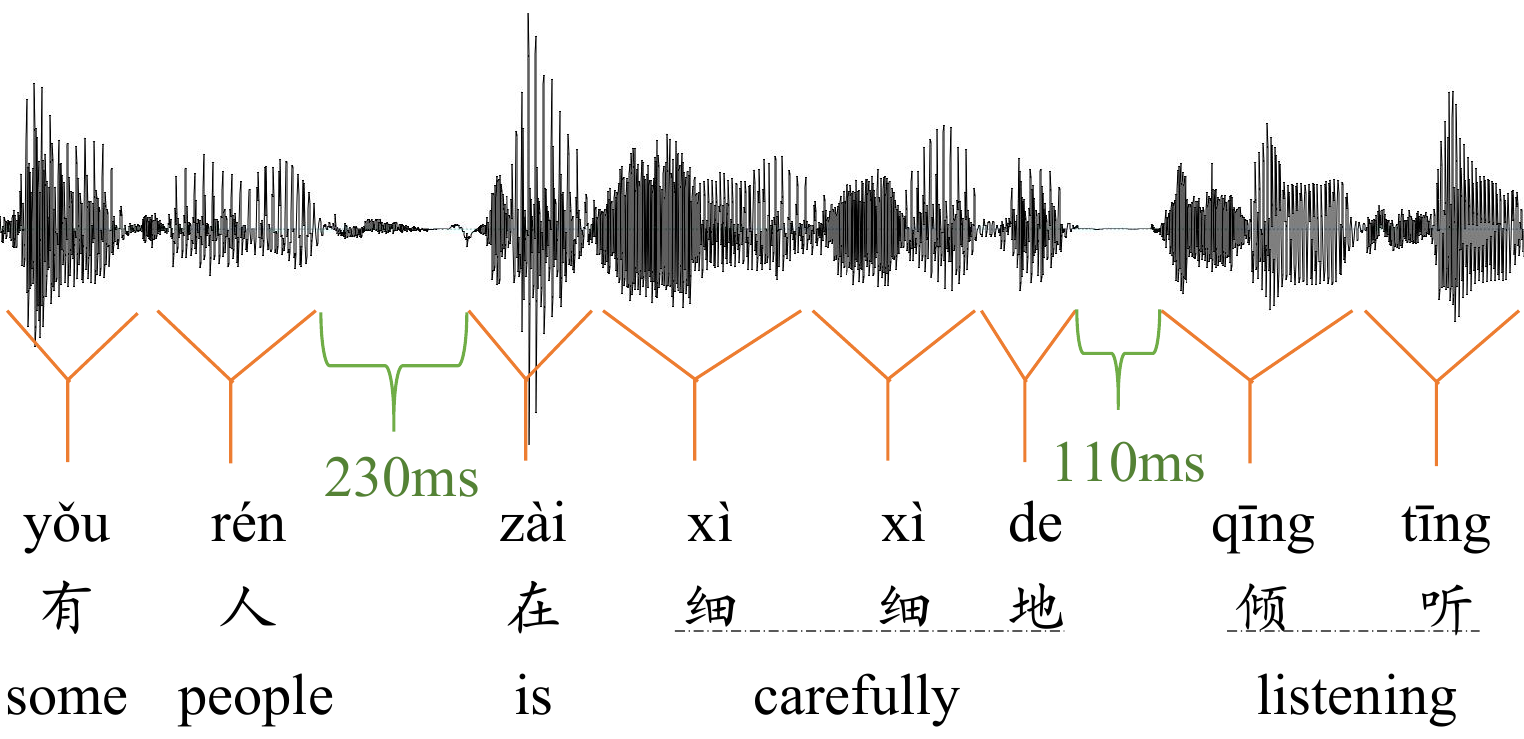}
\caption{An example of speech-text alignment data. The correct segmentation result is  ``有/人/在/细细/地/倾听'', translated as ``some people is carefully listening''. 
}
\label{img:example_extract_pause}
\end{figure}

As a fundamental task in Chinese language processing, CWS aims to segment an input character sequence into a word sequence, since words, instead of characters, are the basic meaning unit in Chinese.  
Figure \ref{img:example_extract_pause} gives an example of the CWS task, along with the speech signals.

With the rapid progress of deep learning techniques, especially the proposal of pre-trained language models like BERT~\cite{devlin-etal-2019-bert}, CWS models have achieve very high performance when there is abundant training data from the same domain as the test data~\cite{tian2020improving,huang2020towards}. Therefore, recent studies on CWS have increasingly focused on the cross-domain scenarios \cite{huang2020joint,ke2021pre}.

Meanwhile, considering the high cost of manually annotating high-quality CWS data, it has been an attractive research direction to explore naturally annotated CWS data from different channels. 
For instance, anchor texts in HTML-format web documents imply reliable word boundaries~\cite{jiang2013discriminative,yang2014semi}; domain-aware dictionaries can match words accurately in target domain texts~\cite{liu2014domain}. 
These studies illustrate that such information can be used as partial annotations for training CWS models. 


Another interesting research line in recent years is the multi-modal integration of speech and texts, mainly due to the adoption of unified model architectures in both speech processing ~\cite{baevski2020wav2vec,hsu2021hubert} and NLP fields ~\cite{devlin-etal-2019-bert,lewis2020bart} in the deep learning era. 
These approaches can be broadly divided into three categories, i.e., 
1) using speech as extra features for NLP ~\cite{zhang2021more}, 2) multi-task learning (MTL) with cross-attention interaction ~\cite{sui2021large}, and 3) end-to-end language analysis from speech ~\cite{chen2022aishell}.
Among these, a work ~\cite{zhang2021more} is closely related to ours. They extract extra features from speech to enhance CWS on corresponding texts. 

Inspired by the progress of research directions discussed above, we propose for the first time to explicitly utilize pauses in speech as word boundary annotations.  
The basic motivation is that when uttering a Chinese sentence, people often pause after finishing some complete meaning in the middle of the sentence, to breathe or to make the speech easier to understand. Considering that words are the basic meaning unit, we hypothesize that pause information can be utilized to help CWS. 


Following previous works on cross-domain CWS, we employ the Chinese Penn Treebank 5 (CTB5) \citep{xue2005penn} as the source domain and use the widely used ZhuXian (``Jade Dynasty'' in English, abbreviated as ZX) data as the target domain \citep{zhang2014type}. 
We collect and clean the parallel speech-text corpus of ZX for mining word boundaries. 
To more thoroughly evaluate the models, we use AISHELL2 as the second target domain, which is a publicly available dataset for automatic speech recognition (ASR)~\citep{du2018aishell2}. 
The contributions of our work are as follows. 

\begin{itemize}[leftmargin=*]
    \item We have manually annotated about 1,000 sentences as the dev/test evaluation data for the AISHELL2 domain. 
    \item We employ the MFA toolkit\footnote{\url{https://mfa-models.readthedocs.io/en/latest/acoustic}}~\citep{mcauliffe2017montreal} to perform character-level alignment on speech-text corpora, and conduct detailed analysis on the collected pauses.  
    \item We propose an effective probability-based strategy for filtering unreliable word boundaries, and a robust CTT strategy to make use of the word boundaries as naturally annotated data. 
    \item Experiments on both ZX and AISHELL2 demonstrate the effectiveness of our proposed approach. We are currently conducting experiments on a much larger dataset, named the Emilia dataset \cite{he2024emiliaextensivemultilingualdiverse} and will report additional results in the Arxiv version of this paper.
\end{itemize}

Our code and newly annotated data have been released and are available at \href{https://github.com/XuebinWang-ai/Mining_Word_Boundaries}{GitHub}.

Please also note that an early version of this work is reported in the arXiv:2210.17122 paper.

\section{Mining Word Boundaries from Speech}

This section describes how we collect speech pauses from parallel speech-text data, which consists of two steps.
First, we prepare parallel speech-text data.
Second, we utilize a GMM-HMM based model to obtain character-level speech-text alignments. Based on the alignments, we can obtain the pause duration between characters. 
Finally, we conduct detailed analysis on pauses and propose a simple filtering strategy to keep reliable pauses as word boundaries.

\subsection{Preparing Speech-Text Parallel Data}
\label{section_collecting_data}

\begin{table}[tb]
    \centering
    \scalebox{0.785}{
    \setlength{\tabcolsep}{3.5pt}
    \renewcommand{\arraystretch}{1.1}
    \begin{tabular}{llrrr}
    \toprule 
    Corpus & Item & Train & Dev & Test\\
    \midrule
    \multirow{2}*{CTB5} & \# Sent & 18,104 & 352 & 348 \\
    ~ & \# Word & 493,932 & 6,821 & 8,008 \\
    \midrule
    \multirow{2}*{ZX} & \# Sent & ~ & 788 & 1,394 \\
    ~ & \# Word & ~ & 20,393 & 34,355 \\
    \midrule
    \multirow{1}*{AISHELL2} & \# Sent &  ~  &  306  & 643 \\
    \multirow{1}*{(Annotated)} & \# Word & ~ & 2,125 & 4,366 \\
    \midrule
    \midrule
    \multicolumn{2}{l}{Speech-text Data} & ~ & \# Pause & \# Sent \\
    \midrule
    \multirow{4}*{ZX} & all & ~ & \multicolumn{1}{c}{---} & 25,038 \\
    ~ & \multicolumn{2}{l}{containing pause}  & 203,842 & 25,016 \\
    ~ & \multicolumn{2}{l}{after filtering (${p}^\texttt{B} \geq 0.1$)} & 198,361 & 25,007 \\
    ~ & \multicolumn{2}{l}{after filtering (${p}^\texttt{B} \geq 0.5$)} & 197,981 & 24,997 \\
    ~ & \multicolumn{2}{l}{after filtering (${p}^\texttt{B} \geq 0.9$)} & 197,540 & 24,964 \\
    
    \midrule
    \multirow{4}*{AISHELL2} & all & ~ & \multicolumn{1}{c}{---} & 847,662 \\
    ~ & \multicolumn{2}{l}{containing pause}  & 537,986 & 324,577 \\
    ~ & \multicolumn{2}{l}{after filtering (${p}^\texttt{B} \geq 0.1$)}  & 457,007 & 294,694 \\
    ~ & \multicolumn{2}{l}{after filtering (${p}^\texttt{B} \geq 0.5$)}  & 449,458 & 290,319 \\
    ~ & \multicolumn{2}{l}{after filtering (${p}^\texttt{B} \geq 0.9$)}  & 442,633 & 286,608 \\
    \bottomrule 
    \end{tabular}
    }
    \caption{Statistics of data used in our experiments. $p^{\texttt{B}}$ means the probability threshold for filtering pauses.
    }
    \label{tab:data-overall}
\end{table}

In this work, we use CTB as the source domain and employ two target-domain datasets. 
Table \ref{tab:data-overall} shows the data statistics. 

\paragraph{(1) ZX.} 
The first dataset is the ZX dataset for the web fiction domain, which was constructed by \citet{zhang2014type} and has been widely used in previous works on cross-domain word segmentation~\citep{liu2012unsupervised,ding-etal-2020-coupling,jiang-etal-2021-fine}.

The ZX dataset contains about 5K sentences in total.\footnote{Among them, 2,373 sentences are reserved for training, but usually are not used in cross-domain experiments.} 
The ZhuXian fiction consists of about 30K sentences in total. 
In this work, we manage to derive word boundaries from speech for the remaining sentences that are not included in ZX-dev/test. 

We select a version\footnote{\url{https://ting55.com/book/143}} characterized by high quality and little background noise from various iterations available online.
All audios are processed to be at a sampling frequency of 16kHz.  

\paragraph{Cleansing.} We apply several data cleansing or filtering strategies to improve the data quality. (1) Numbers like ``1200'' are transformed into their Chinese character form like ``一千两百'' (one thousand and two hundred). (2) Silent and special symbols in the texts like punctuation marks are removed.
(3) Irrelevant blanks or noises in the beginning or end of the audio are removed. (4) Audios with background music are discarded. 
Finally, we collect 246 audio files amounting to 144 hours, each corresponding to a chapter of the fiction.

\paragraph{(2) AISHELL2.} 
For the second domain, we adopt the AISHELL2~\citep{du2018aishell2} Mandarin Chinese speech corpus, which contains about 1,000 hours of high-quality audio, corresponding to about one million transcription sentences.\footnote{We sincerely thank the Beijing AISHELL Technology Co., Ltd for sharing the data. }  
The corpus covers 12 different domains that 
are closely related with 
application of speech recognition in smart home, autonomous driving, industrial production, etc.

One major feature of the AISHELL2 data, whose major use is as training data for ASR, is that the transcription texts do not contain punctuation marks. In fact, outputs of ASR models usually do not contain soundless symbols in written texts, including punctuation marks. 

Instead of injecting punctuation marks into AISHELL2 transcription texts, which would be highly time-consuming and prone to annotation errors, we decide to perform word segmentation on transcription texts directly. We believe this is an interesting and useful scenario for word segmentation research. Text normalization procedures such as filling punctuation marks may be applied over the output word sequence.  

To alleviate the mismatch between the AISHELL2 data and the source-domain training data, i.e., CTB, regarding punctuation marks, we employ a simple strategy that can boost the performance of the baseline model. 
For each sentence in CTB-Train, we remove the punctuation marks in the sentence. 
With this strategy, the trained model can handle transcription texts well. 

To evaluate the CWS model on AISHELL2, we have manually annotated about 1,000 sentences in the original AISHELL2-dev/test, and use them as  the dev/test evaluation datasets. 
We present  more details about data annotation in Section 
\ref{sec:annotation-detail-aishell2}.

\subsection{Character-level Speech-Text Alignment}
\label{sec:speech_text_alignment}
In this paper, we try to derive word boundaries from speech based on pause information. 
The intuition is that if the speaker pauses for some time after uttering a character, then there may be a word boundary after the character. 
The key challenge for implementing this idea is how to obtain accurate character-level alignments between speech signals and the corresponding sentence. 

In the past decade, end-to-end Transformer based models have become the dominant ASR approach due to its superior performance~\citep{Gulati2020,Zhang2023GoogleUS,pratap2023scaling}. 
With an extra Connectionist Temporal Classiﬁcation (CTC) component, the model can explicitly produce alignments. However, our early experiments reveal that the Transformer-CTC based models suffer from a severe peak alignment issue, meaning that every character is usually aligned to a single speech frame, leaving most of the frames aligned to blanks. 
This finding is consistent with previous results~\citep{senior2015acoustic,zeyer2021does}. 

Instead, we employ the MFA toolkit with its GMM-HMM implementation to obtain alignment between text and speech \citep{mcauliffe2017montreal}. 
We employ both monophone and triphone GMMs. 

Given a speech, we use the default frame window length of 25ms and the default frame offset of 10ms. 
For each frame, the acoustic features are the standard Mel-Frequency Cepstral Coefficients (MFCCs). 
Formally, we represent speech as $\mathbf{x}=x_0...x_i...x_n$, where $x_i$ is an MFCC feature vector, and the corresponding transcription as $\mathbf{y}=y_0...y_i...y_{m}$, where $y_i$ denotes a token. 
The objective of GMM-HMM is two fold: 1) to determine which phonemes correspond to a token, and 2) to determine which frames (e.g., $x_k...x_l$) correspond to a phoneme. 
Combining the results, we can obtain the 
time range for each token. 
The model works under the unsupervised scenario, and apply the expectation-maximization (EM) algorithm~\citep{moon1996expectation} on the training speech-text pairs. 

We continue training the pre-trained Mandarin model in the MFA toolkit using our parallel speech-text data at hand, either ZX or AISHELL2. 
In our context, a token $y_i$ corresponds to a character.\footnote{
By default, the Mandarin model in the MFA toolkit can only perform alignment at the word level, since the acoustic dictionary is word-based and polyphonic characters only have one entry, corresponding to the most frequent pronunciation. To handle this issue, we extend the acoustic dictionary by leveraging a Pinyin-based Chinese lexicon (both words and characters). We will release the related resource and scripts.  
} 
Suppose $y_i$ is aligned to $x_{b_i}...x_{e_i}$, also denoted as $(b_i, e_i)$, where $b_i$ and $e_i$ are the beginning and end indices of frames. 
Then we can calculate the pause duration between two adjacent characters, for instance $y_i$ and $y_{i+1}$ as follows.
\begin{equation}
d(y_i,y_{i+1}) = (b_{i+1} - e_i) \times \textit{10ms} 
\label{formula:char_interval}
\end{equation}
Figure \ref{img:example_extract_pause} gives an example. 
There are two pauses in the sentence, with duration of $\textit{230ms}$ and $\textit{110ms}$ respectively.

\subsection{Filtering Pauses} 
\label{sec:filtering-pauses}
At the beginning, our plan was to filter unreliable word boundaries based on a global pause duration threshold. 
For instance, if $d(y_i,y_{i+1}) < \textit{50ms}$, then we discard the pause and do not consider it as a boundary. 
However, our analysis shows that pauses with short duration are equally helpful. 

Then we turn to another simple probability-based filtering strategy. The idea is to let the baseline model trained on the source-domain data (i.e., CTB) to judge. 
If the baseline shows a low probability for a boundary, we discard it.

Following previous works, we adopt the BERT-CRF model as our baseline model and employ the label set $\texttt{\{B, M, E, S\}}$, which represents “beginning”, “middle”, “end”, and “single-char”, respectively. 
Given an input char sequence $\mathbf{y}=y_0...y_{m}$, we denote a label sequence as $\mathbf{z}=z_0...z_{m}$. 
The marginal probability of a label bigram at given positions $i$ and $i+1$, for instance $\texttt{E\_S}$, is:
\begin{equation}
p(\texttt{E\_S} |\mathbf{y}, i) = \sum_{\mathbf{z}:z_i=\texttt{E}, z_{i+1}=\texttt{S}}p(\mathbf{z} | \mathbf{y}).
\end{equation}

Then the probability that there is a boundary between $y_i$ and $y_{i+1}$ is: 
\begin{equation}
p^\texttt{B}(\mathbf{y}, i)  = \sum_{l \in \{ \texttt{S\_S}, ~ \texttt{S\_B}, ~ \texttt{E\_S}, ~ \texttt{E\_B} \}}p(l |\mathbf{y}, i).
\end{equation}
And the probability that there is no boundary is:
\begin{equation}
1- p^\texttt{B}(\mathbf{y}, i)  = \sum_{l \in \{ \texttt{B\_M}, ~ \texttt{B\_E}, ~ \texttt{M\_M}, ~ \texttt{M\_E} \}}p(l |\mathbf{y}, i).
\end{equation}
Please note that illegal label bigrams (a.k.a. illegal transitions), such as \texttt{B\_B}, are forbidden and always receive zero probability.

According to our experiments and analysis, our final approach keeps all pauses with $p^\texttt{B} \geq 0.5$, regardless of the pause duration.

\subsection{Analysis of Pauses} 
\label{sec:further-analysis-pause}

\begin{figure}[tb]
\centering
\includegraphics[width=7.5cm]{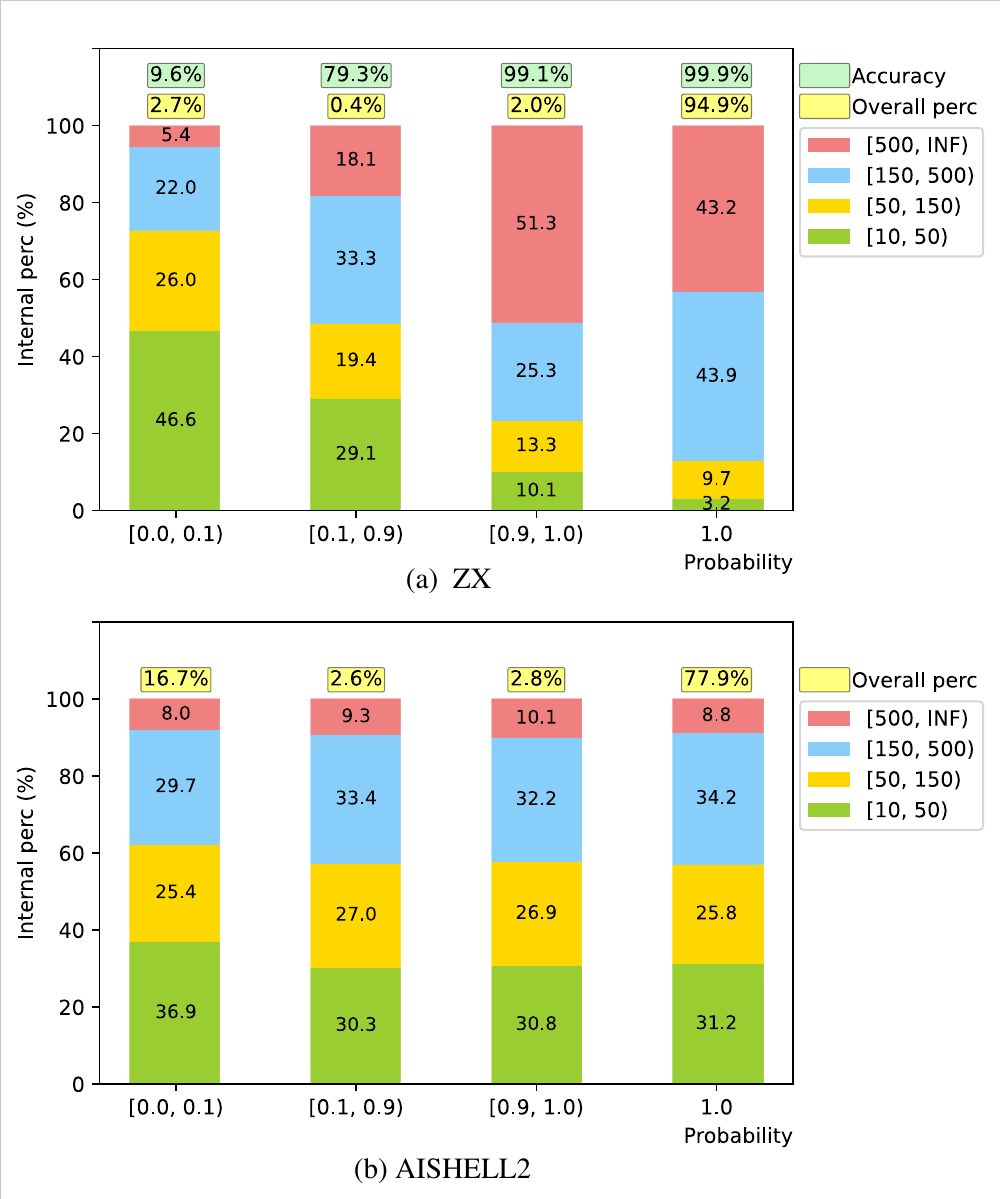}
\caption{Statistics of pauses regarding probability/accuracy of being boundaries and duration distribution. 
Probabilities are grouped into four bins, i.e., ${[0.0, 0.1)}$, ${[0.1, 0.9)}$, ${[0.9, 1.0)}$, and ${1.0}$. 
The overall \uline{perc}entage means the proportion of pauses belonging to a given probability bin against all pauses. 
Pause durations are divided into four bins, i.e., ${[10,50)}$, ${[50,150)}$, ${[150,500)}$, and ${[500,\texttt{INF})}$, in the unit of $\textit{ms}$. 
Given a probability bin, the internal \uline{perc}entage 
means the proportion of pauses belonging to a given duration bin against all pauses in the probability bin. 
For the ZX data, accuracy means the proportion of pauses that are really word boundaries according to further verification. 
}
\label{pauses}
\end{figure}

The lower part of Table \ref{tab:data-overall} presents the overall statistics of pauses in both ZX and AISHELL2, both with and without filtering. One notable difference between the two datasets is that pauses are much sparser in the latter. 
Almost all sentences in ZX contain pauses ($\geq 10ms$), and for sentences that contain pauses, the average number of pauses is about $8$. 
In contrast, less than 40\% of sentences in AISHELL2 contain pauses, and the average number is only $1.7$. 
We believe that the major reason is that the sentences are much longer in ZX than in AISHELL2. Each sentence contains about 25 words on average in the former, while only about 7 in the latter. 

Figure~\ref{pauses} provides more details about the pauses. We group probability of $[0.1,0.9)$ into one bin for two reasons. First, the total percentage of pauses falling into this bin is still not high. Second, pauses within the bin scatter quite evenly in terms of probability. Our experiments show that despite the low overall percentage, pauses in this bin are quite valuable for improving model performance. 

From the aspect of \emph{overall percentage}, the most notable difference is that the percentages for the first two probability bins, i.e., $[0.0,0.1)$ and $[0.1,0.9)$, are much higher in AISHELL2 than in ZX ($2.7 \rightarrow 16.7$ and $0.4 \rightarrow 2.6$). 

From the aspect of \emph{internal percentage}, we can see that pauses of different duration bins have a similar distribution in the four probability bins in AISHELL2. 
In contrast, in ZX the percentages of smaller pause durations, i.e., $[10,50)$ and $[50,150)$, decrease consistently as the probability increases. 

For ZX, we also manage to report the accuracy for each probability bin, in order to gain more insights. 
Instead of performing manual annotation, we notice that the ZX data with word segmentation (WS) annotations are a part of the transcription texts. Thus, we evaluate the accuracy of pauses as word boundaries over the overlapping sentences, using annotated WS information as gold standard.\footnote{
Due to several factors, including transcription mistakes, 
difference in the fiction versions,  
difference in sentence segmentation procedures, etc, we collect about 2K overlapping sentences that appear both in the transcription texts and the ZX data. 
} 

It is clear that accuracy increases consistently as the probability becomes higher. Most of the pauses falling into the $[0.0, 0.1)$ bin are incorrect boundaries and thus should be excluded. 

Pauses with high probability, i.e., $[0.9, 1.0)$ and $1.0$, have almost perfect accuracy and should be included. 
Despite the model has high confidence in these word boundaries, they are valuable additions to the cross-domain training dataset due to the extensive data volume.

Most importantly, pauses in the $[0.1, 0.9)$ have $79.3\%$ accuracy, which is much higher than that for the $[0.0, 0.1)$ bin. Our experiments show that these pauses are very useful for the model.

\section{Utilizing Pauses as Word Boundaries}

\label{section_comeplete_then_train}

\paragraph{Pauses as word boundaries for CWS.}
In fact, quite a few previous studies try to explore word boundaries from different channels and use them as naturally annotated CWS data \cite{jiang2013discriminative,liu2014domain,yang2014semi}. 
Under a sequence labeling framework, word boundaries can be naturally treated as partial annotations and used to construct a constrained label space.
A constrained label space refers to a set of allowed labels for each character in a sequence, based on the identified word boundaries. This space restricts the possible labels for each character, and similarly, labels that contradict existing annotations are excluded. The constrained label space ensures that only correct word boundaries are considered.

\begin{figure}[tb]
\centering
\includegraphics[width=7.5cm]{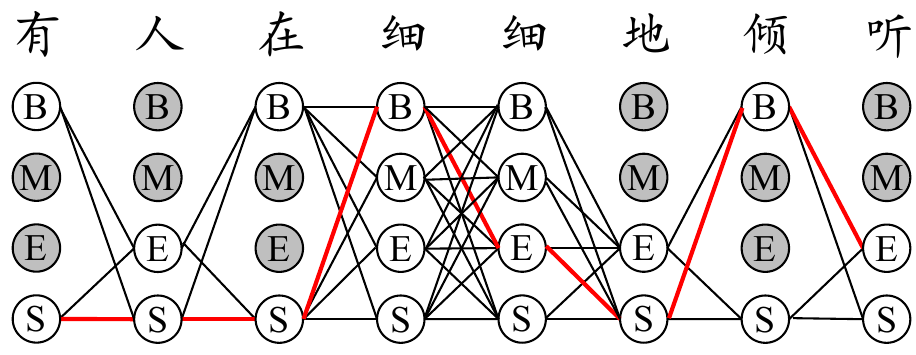}
\caption{Constrained label space for the sentence in Figure~\ref{img:example_extract_pause}, in which we obtain two boundaries ``有人/在细细地/倾听''. 
Illegal labels are marked as gray. The red thick lines present a legal path. 
In this context, the character ``人 (people)'' is constrained to be either a single-char word (``S'') or the end of a word (``E'') due to the pause after it. This constraint is based on the assumption that the pause indicates a clear word boundary, preventing ``人'' from being labeled as the beginning (``B'') or middle (``M'') of a multi-char word.
}
\label{img:natural_annotation_constraint}
\end{figure}

Figure~\ref{img:natural_annotation_constraint} gives an example. Due to the pause ``人 (people) / 在 (is)'', the left-side char ``人'' can only be either a single-char word (``S'') or the end of a word (``E''), while the right-side char ``在'' can only be either a single-char word (``S'') or the beginning of a word (``B''). 
A similar explanation goes to the second boundary. 

\subsection{Problem with the Partial-CRF strategy}

To make use of partially annotated training samples, shown in Figure~\ref{img:natural_annotation_constraint}, 
we first employ the partial-CRF strategy \citep{liu2014domain}, which is theoretically elegant.
The basic idea is that 
instead of maximizing the probability of a single gold-standard label sequence, the training objective is to maximize the sum of probabilities of all legal paths in the constrained space, which can be efficiently computed via a variant of the Forward algorithm. 

However, our experiments show that this strategy performs terribly when the model is trained on both CTB-Train and the target-domain data with partial boundaries. 
Further analysis shows the model heavily predicts the  ``$\texttt{S}$'' label for target-domain sentences (i.e., most words being single-char).
We suspect the major reason is that 
all characters in the constrained space can be labeled as ``$\texttt{S}$'' tags, as shown in Figure~\ref{img:natural_annotation_constraint}, and the model fails to transfer from CTB to the target domain the knowledge of when/how to compose multi-char words. 

\begin{figure}[tb]
\centering
\includegraphics[width=7.5cm]{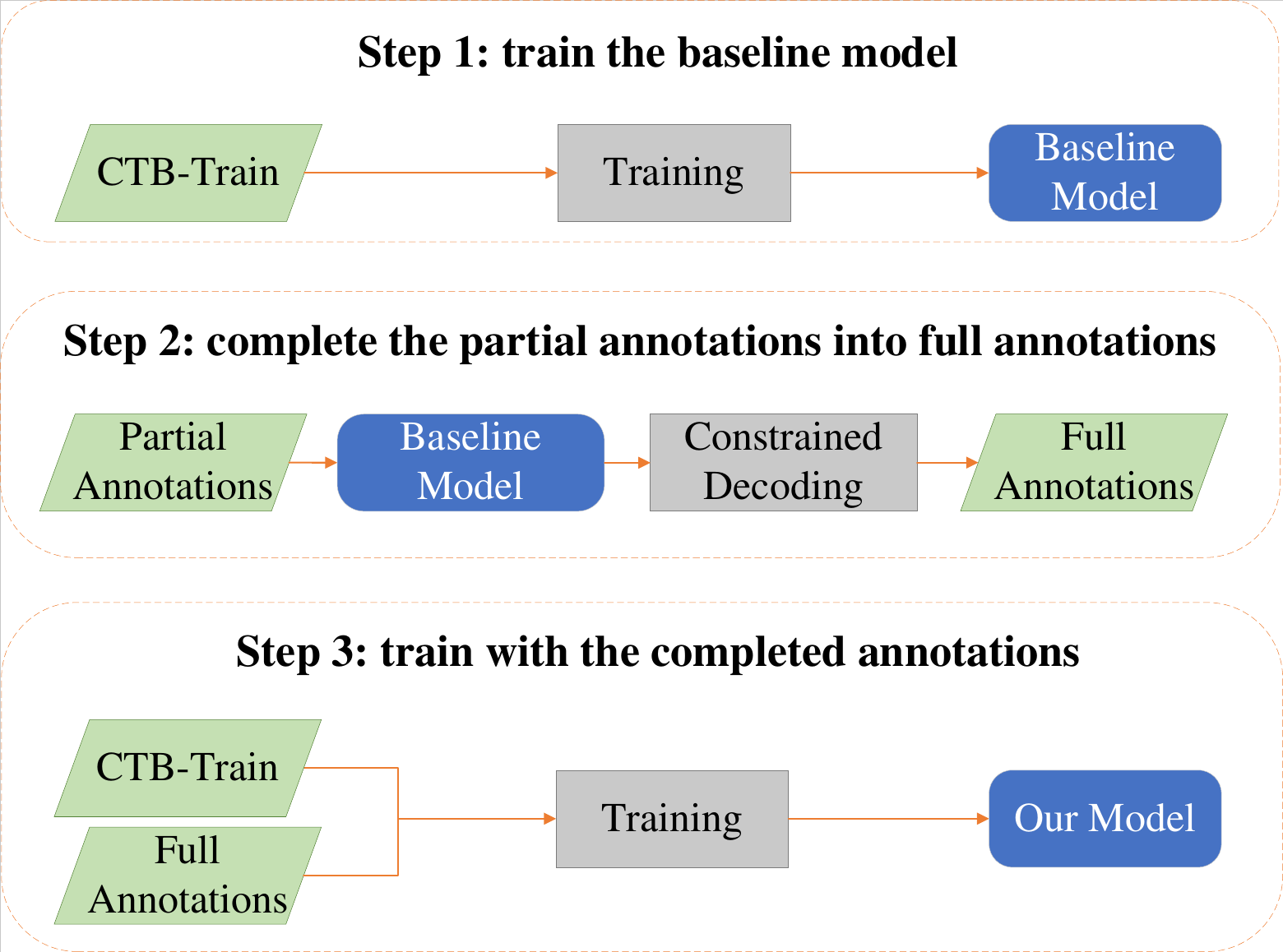}
\caption{The CTT training strategy.}

\label{img_complete_then_train}
\end{figure}

\subsection{The Complete-Then-Train (CTT) Strategy}

To address the above issue, we present a simple yet effective CTT strategy. 
The idea is converting partial annotations into full annotations by letting a basic model select an optimal sequence in the constrained space. 
Figure~\ref{img_complete_then_train} illustrates the strategy, consisting of three steps. 
First, we train a CWS model (i.e., baseline) on the source-domain dataset. 
Second, we employ this baseline to complete partial annotations into full ones. More concretely, the baseline selects an optimal label sequence through constrained Viterbi decoding. 
For example, the model chooses the red thick lines path in Figure~\ref{img:natural_annotation_constraint}. 
Lastly, we use both source-domain and completed data to train the full model.

\section{Experiments}
\label{sec:experiments}

\subsection{Annotation Details for AISHELL2}

\label{sec:annotation-detail-aishell2}
Upon release, the AISHELL2 dataset set aside 2,500 sentences and 3,000 sentences, serving as the dev and test sets, respectively. 
We apply the baseline models and our full models to the 5,500 sentences. 
From the sentences that receive different results from a baseline model and a full model, we collect about 1,000 sentences for manual annotation.

Two postgraduate students participate in the data annotation. 
Our annotation process consists of two stages. 
At the first stage, each sentence is annotated by two annotators, and the differences are resolved by further discussion. 
During this stage, the annotators become familiar with the segmentation guidelines of CTB~\cite{xia2000segmentation}. 
At the second stage, one annotator (the first author of this submission) conducts a thorough review and correction of the annotations. 
We plan to annotate additional sentences to make the experimental conclusions more solid.

To speed up the annotation process, we provide the results of the two models with differences highlighted.
Meanwhile, the model outputs are randomized to ensure annotators cannot tell which results are from which model, thereby avoiding any bias towards our method. 
Table \ref{tbl:anno_example} illustrates the annotation process.

After removing sentences that cannot be labeled due to noise or transcription errors, we obtain 949 sentences in total. 
We reorganize them into new dev and test sets based on their original set affiliations (dev or test).
Table \ref{tab:data-overall} shows the data statistics. 

\begin{table}[tb]
    \centering

    \begin{small}
    \setlength{\tabcolsep}{3.5pt}
    \renewcommand{\arraystretch}{1.1}
    \begin{tabular}{ll}
    \toprule 
    Item & Sentence \\
    \midrule
    \multirow{2}*{} & 邀请上朋友办个晚宴 \\
    ~ & Invite friends to host a dinner party \\
    \midrule
    \multirow{2}*{Results of Model 1} & 邀* / 请上* / 朋友 / 办 / 个 / 晚宴 /  \\
    ~ & Invite / please up / friends / to host / \\
    ~ & a / dinner party \\
    \midrule
    \multirow{2}*{Results of Model 2} & 邀* / 请* / 上* / 朋友 / 办 / 个 / 晚宴 / \\
    ~ & Invite / please / up / friends / to host / \\
    ~ & a / dinner party \\
    \midrule
    \multirow{2}*{Annotation Results} & 邀  请 / 上 / 朋  友 / 办 / 个 / 晚  宴 / \\
    ~ & Invite / friends / to host / a / dinner party \\
    \bottomrule 
    \end{tabular}
    \end{small}
    \caption{Illustration of the annotation process of the AISHELL2 dev/test data. * highlights differences in model results. 
    }
    \label{tbl:anno_example}
     
\end{table}

\subsection{Settings}


For the evaluation, we employ the standard metrics of precision (P), recall (R), and the F1 score.

As discussed in Section \ref{sec:filtering-pauses}, we regard CWS as a sequence labeling task and employ the BERT-based\footnote{\url{https://huggingface.co/bert-base-chinese}} CRF baseline model. 
We use AdamW with an initial learning rate of 5e-5, and a mini-batch size of 1000 characters. 
The dropout ratio is 0.1 for all models. 
We train each model for 10 epochs. 

Following previous works on cross-domain WS on ZX, we use CTB5-Train and full annotations as the training data, and use the target-domain dev set to select the best iteration. 

To be more convincing, we train each model three times with three different random seeds and present the average and standard deviation.\footnote{$\sigma=\sqrt{\frac{1}{n-1}{\sum_{k=1}^n(x_i-\bar{x})^2}}$}

\begin{table*}[tb]
    \centering
    \begin{small}
    
    \setlength{\tabcolsep}{3.5pt}
    \renewcommand{\arraystretch}{1.0}
    \scalebox{0.95}{
    \begin{tabular}{lccccccc}
    \toprule 
    ~ & P & R & F1 & P & R & F1\\
    \midrule
    Models & \multicolumn{3}{c}{ZX-dev} & \multicolumn{3}{c}{ZX-test} \\
    
    \midrule
    Baseline & 94.16 & 94.39 & 94.27$_{\pm0.20}$ & 93.16 & 93.82 & 93.49$_{\pm0.22}$ \\    
    Using word boundaries & ~ & ~ & ~ & ~ & ~ & ~  \\
    \quad w/o filtering & 94.18 & 94.34 & 94.26$_{\pm0.24}$ & 93.69 & 94.03 & 93.86$_{\pm0.36}$ \\

    \quad w/ filtering (${p}^\texttt{B} \geq 0.9$), self-training
    & 94.27 & 94.64 & 94.45$_{\pm0.17}$ & 93.46 & 94.08 & {93.77}$_{\pm0.25}$ \\

    \quad w/ filtering (${p}^\texttt{B} \geq 0.5$)
    & 94.33 & 94.66 & \textbf{94.56}$_{\pm0.39}$ & 93.59 & 94.22 & {93.90}$_{\pm0.30}$ \\
    
    \quad w/ filtering (${p}^\texttt{B} \geq 0.1$) & 94.23 & 94.78 & 94.50$_{\pm0.27}$ & 93.56 & 94.32 & \textbf{93.94}$_{\pm0.20}$ \\
    
    \midrule
    \multicolumn{7}{c}{Previous Results}  \\
    \midrule
    \citet{ding-etal-2020-coupling}  & ~ & ~ & --- & ~ & ~ &  90.90~~~~~~~~ \\
    \citet{luo-etal-2022-ji} & ~ & ~ & --- &  & ~ & 91.11~~~~~~~~ \\
    \citet{ShoheiHigashiyama2020}  & ~ & ~ & --- & ~ & ~ & 93.30~~~~~~~~ \\
    
    \midrule
    \midrule

    Models & \multicolumn{3}{c}{AISHELL2-dev} & \multicolumn{3}{c}{AISHELL2-test} \\
    \midrule
    Baseline & 89.08 & 90.10 & 89.58$_{\pm0.48}$ & 88.20 & 88.43 & 88.31$_{\pm0.34}$ \\
    Using word boundaries & ~ & ~ & ~ & ~ & ~ & ~ \\
    \quad w/o filtering & 89.32 & 90.81 & 90.06$_{\pm0.52}$ & 87.88 & 88.76 & 88.31$_{\pm0.15}$  \\
    \quad w/ filtering (${p}^\texttt{B} \geq 0.9$), self-training
    & 90.59 & 90.89 & 90.74$_{\pm0.21}$ & 88.39 & 88.36 & 88.38$_{\pm0.41}$ \\ 
    
    \quad w/ filtering (${p}^\texttt{B} \geq 0.5$) & 90.82 & 90.84 & 90.83$_{\pm0.58}$ & 89.45 & 88.63 & \textbf{89.04}$_{\pm0.26}$ \\

    \quad w/ filtering (${p}^\texttt{B} \geq 0.1$) & 90.65 & 91.06 & \textbf{90.85}$_{\pm0.47}$ & 89.02 & 88.67 & 88.84$_{\pm0.25}$ \\

    \bottomrule 
    \end{tabular}
    }
    \end{small}
    \caption{Main results on both datasets. 
    }
    \label{tab:cross_domain}
    
\end{table*}

\subsection{Results}

Table~\ref{tab:cross_domain} presents the main results. Compared with previous results on ZX, our baseline models already achieve very good performance. 

Most importantly, 
our best models using filtered pauses as word boundaries achieve significant improvement of 0.45 and 0.73 in F1 score on ZX-test and AISHELL2-test, respectively, compared with the baseline models. 

\paragraph{Effect of filtering pauses.} In comparison to models without filtering pauses, our final models (${p}^\texttt{B} \geq 0.5$) are consistently superior in F1 scores. 

To enhance comprehension of our approach, we train the CWS model on datasets without speech information, 
i.e., we directly complete target-domain annotations instead of constrained decoding. 
This technique, referred to as self-training~\cite{zhou-etal-2024-chinese}, aligns with our approach using word boundaries with ${p}^\texttt{B} \geq 0.9$, as evidenced by the experimental results in Table~\ref{tab:cross_domain}.

\paragraph{Usefulness of word boundaries with probability of $[0.1,0.9)$.}  On the one hand, compared with using the self-training method, our final models are consistently superior in F1 scores. 
On the other hand, compared with baselines, models using all pauses have even lower F1 scores on ZX-test and AISHELL2-test.
These two aspects highlight the effectiveness of pauses with probability of $[0.1,0.9)$. 

To better explore the role of word boundaries within the probability interval $[0.1, 0.9)$ on the model, we take the AISHELL2 dataset, which has a larger number of word boundaries than ZX, as an example to conduct more detailed experiments. 
The results presented in Table~\ref{tab:AISHELL2_1_9} indicate that word boundaries within the range $[0.5, 0.9)$ have the most positive impact on the model performance.
The utilization of word boundaries with probabilities falling within the interval $[0.1, 0.5)$ has resulted in a detrimental impact on performance.
In conjunction with Table~\ref{tab:cross_domain}, our analysis reveals that word boundaries within the probability range $[0.1, 0.5)$ exhibit only slight negative effects when trained on entire datasets.

\begin{table}[b]
    \centering
    \begin{small}
    \setlength{\tabcolsep}{3.5pt}
    \renewcommand{\arraystretch}{1.1}
    \begin{tabular}{lcc}
    \toprule 
    Models  & AISHELL2-test-F1 \\
    \midrule
    Word boundaries ($0.1 \leq {p}^\texttt{B} < 0.5$) \\
    \quad Self-training & \textbf{88.19}  \\
    \quad Our method & 87.20 \\    
    \midrule
    Word boundaries ($0.5 \leq {p}^\texttt{B} < 0.9$) \\
    \quad Self-training & 88.14  \\
    \quad Our method & \textbf{88.33} \\
    
    \bottomrule 
    \end{tabular}
    \end{small}
    \caption{Comparative experiments on AISHELL2. 
    }
    \label{tab:AISHELL2_1_9}
\end{table}

\subsection{Additional Results on Larger Datasets}

As illustrated in Figure~\ref{pauses}, the quantity of effective pauses is limited due to the small size of the dataset we used. 
Therefore, we plan to conduct experiments on the Emilia dataset~\citep{he2024emiliaextensivemultilingualdiverse} to mine more word boundaries.
However, given the substantial volume of data and the time constraints, we have yet to complete this experiment. We will provide updates on ArXiv upon completion.

\section{Related Works}
\label{sec:related_works}

\subsection{Integrated Speech and Text Processing} 

In deep learning, the Transformer-based model architecture becomes popular in both speech processing and NLP fields. 
The same architecture makes it convenient to process speech and textual data in an integrated manner. Intuitively, speech and text can provide complementary useful features. We summarize recent works into four groups.  

\paragraph{(1) Speech as extra features for NLP.}
The most straightforward way is to extract features from speech and use them as extra inputs for an NLP model. \citet{zhang2021more} make a pioneer effort to use speech features for CWS, which is closely with our work. Their approach requires parallel speech-text data in both training and test phases, with WS annotations and the character-frame alignments. They manually annotate 250 sentences and split them into training-test data. Experiments show that extra speech features are beneficial. 

Different from their work, ours emphasis on the use of pause information in speech. We do not need WS annotations for the text data and automatically derive character level alignments. In the test phase, our CWS model performs only on text data, rather than parallel speech-text data. 

\paragraph{(2) MTL 
with cross-attention interaction.} 
Given speech-text parallel data, \citet{sui2021large} present a multi-task learning approach that performs NER and ASR at the same time. They first use separate encoders for the two types of inputs, and then employ the cross-attention mechanism to achieve multi-model interaction. 

\paragraph{(3) End-to-End language analysis from speech.}
Several works propose to directly derive language analysis results from speech inputs in an end-to-end manner. 
\citet{ghannay2018end} embed named entity labels into texts and train a model that transcribes speech into texts and treats named entity labels as normal tokens. They conduct experiments on French NER. 
\citet{yadav2020end} apply the approach to English NER and propose a new label embedding scheme. 
\citet{chen2022aishell} present a Chinese datasets of parallel speech-text data with NE annotations, and 
systematically compare the pipeline and end-to-end approaches.

\citet{wu2022towards} propose an end-to-end relation extraction model that transcribes speech into (entity, entity, relation) triples, and  totally ignores the full text (not performing ASR). However, their experiments show that the end-to-end approach is inferior to the pipeline model, i.e., first ASR and then relation extraction on texts.


\paragraph{(4) Utilizing speech pauses.}~\citet{fleck2008lexicalized} utilize speech pauses to aid in word segmentation from transcribed adult conversations. 
Specifically, the pauses serve to bootstrap a discriminative model that determines word boundaries by examining phone ngrams observed before and after pauses.
The algorithm segments the phoneme sequence into words by estimating the likelihood of a phone sequence occurring at the end of a phrase, which is a strong indicator of a word boundary. 
This approach is effective in handling morphologically complex languages like English and Arabic.

\subsection{Cross-domain CWS.} 
\citet{ding-etal-2020-coupling} design a distant annotation method to annotate the target domain text and use the adversarial training strategy to train the cross-domain model. \citet{luo-etal-2022-ji} propose supervised CRF and semi-CRF to train models in both the source and target domains; \citet{ShoheiHigashiyama2020} training bilstm-Affine predicts BMES tags separately to achieve lexicon words prediction. Compared with their method, our method uses the information from speech to annotate the text, and leverage basic model to complete the partial annotated data for further training.

\subsection{Naturally annotated CWS data}

\paragraph{Mining naturally annotated data.}
Previous studies try to mine naturally annotated CWS data from different channels. 
~\citet{jiang2013discriminative} hypothesize that anchor texts (i.e., for hyperlinks) in HTML-format web documents are very likely to correspond to complete meaning units, and thus can be explored to obtain at least two word boundaries. 
In the cross-domain scenario, ~\citet{liu2014domain} use a domain-related dictionary and perform maximum matching on unlabeled target-domain text, treating matched texts as annotated words. 

\paragraph{Utilizing naturally annotated data.} 
Above naturally annotated data are in two forms. In the first form, some word boundaries in the sentence are given, whereas in the second, some words are given. 
Both forms can be treated as partial annotations, in contrast to full annotations, and be encoded as constrained label space as shown in Figure ~\ref{img:natural_annotation_constraint}.

~\citet{jiang2013discriminative} proposes a constrained decoding approach to learn from partially annotated data with word boundaries. 
They use a max-margin training loss. For each training sentence, they first obtain an optimal label sequence from the constrained space and use it as gold-standard reference in an online fashion. 

Some researchers employ the CRF ~\citep{liu2014domain,yang2014semi} to extend the loss for learning from partial/incomplete annotations.
In this work, we also use this approach, but obtain inferior performance probably due to the issue of pervasive ``S'' labels. 
Therefore, we propose a simple yet effective CTT strategy. 

\section{Conclusion}

\label{sec:conclusion}

This paper for the first time proposes to explicitly mine word boundaries from speech-text data as extra naturally annotated training data for cross-domain CWS. 

Firstly, we collect speech-text data from the web fiction domain (ZX) and annotate part of AISHELL2-dev/test datasets for CWS evaluation. 
Secondly, we perform character-level alignment on the speech-text data to mine word boundaries. 
Thirdly, we employ the baseline to calculate the marginal probability of word boundaries. 
By analyzing the accuracy across four probability ranges, we filter out word boundaries with lower probabilities.
Finally, we apply the CTT method to effectively leverage the filtered word boundaries for the annotation of target-domain training data, thereby substantially enhancing the performance of the CWS model in cross-domain settings.
Our experiments demonstrate that mined word boundaries significantly improve CWS via the CTT method. 
Analysis reveals that filtering boundaries is crucial to the efficacy of the CTT method.

\section*{Limitations}
We believe our work has built a solid foundation for future research in this direction. Meanwhile, we are aware that 
our work is limited and can be improved in several aspects. 

First, our approach relies on accurate character-level alignment between speech and texts. 
So far, we have used MFA as a black-box and our early trials showed that the end-to-end Transformer-CTC model is inferior. Therefore, our proposed approach may be more effective with improved alignment quality.  

Second, this work only utilizes pauses detected by character-level aligner to derive word boundaries, but it ignores other rich features in speech. For example, intonation or pitch change may also be helpful. 

Finally, as discussed in \ref{sec:annotation-detail-aishell2}, we plan to annotate more evaluation data for AISHELL2 to make the experiments more solid.





\section*{Acknowledgements}

We thank the anonymous reviewers for their helpful comments and insights regarding our work. 
This work was supported by the National Natural Science Foundation of China (Grant NO. 62176173 and 62306202), and a project funded by the Priority Academic Program Development (PAPD) of Jiangsu Higher Education Institutions.

\bibliography{anthology,custom}


\end{CJK}
\end{document}